\documentclass[letterpaper]{article} 
\usepackage{aaai24}  
\usepackage{times}  
\usepackage{helvet}  
\usepackage{courier}  
\usepackage[hyphens]{url}  
\usepackage{graphicx} 
\usepackage{natbib}  
\usepackage{caption} 
\usepackage{algorithm}
\usepackage{algorithmic}

\usepackage{newfloat}
\usepackage{listings}

\usepackage{booktabs}
\usepackage{tabularx}
\usepackage{nicematrix}
\usepackage[capitalize]{cleveref}

\DeclareCaptionStyle{ruled}{labelfont=normalfont,labelsep=colon,strut=off} 
\floatstyle{ruled}
\newfloat{listing}{tb}{lst}{}
\floatname{listing}{Listing}
\title{A Comprehensive Augmentation Framework for Anomaly Detection}
\author{
	Jiang Lin,Yaping Yan\thanks{Corresponding author}
}
\affiliations{

    School of Computer Science and Engineering, Southeast University, Nanjing 210096, China\\
    Key Laboratory of New Generation Artificial Intelligence Technology and Its Interdisciplinary Applications (Southeast University), Ministry of Education, China\\
    \{220215663, yan\}@seu.edu.cn
}

\usepackage{bibentry}

\begin{document}

\maketitle

\begin{abstract}
Data augmentation methods are commonly integrated into the training of anomaly detection models.
Previous approaches have primarily focused on replicating real-world anomalies or enhancing diversity, without considering that the standard of anomaly varies across different classes, potentially leading to a biased training distribution.
This paper analyzes crucial traits of simulated anomalies that contribute to the training of reconstructive networks and condenses them into several methods, thus creating a comprehensive framework by selectively utilizing appropriate combinations.
Furthermore, we integrate this framework with a reconstruction-based approach and concurrently propose a split training strategy that alleviates the overfitting issue while avoiding introducing interference to the reconstruction process.
The evaluations conducted on the MVTec anomaly detection dataset demonstrate that our method outperforms the previous state-of-the-art approach, particularly in terms of object classes.
We also generate a simulated dataset comprising anomalies with diverse characteristics, and experimental results demonstrate that our approach exhibits promising potential for generalizing effectively to various unseen anomalies encountered in real-world scenarios.

\end{abstract}

\section{Introduction}
Surface anomaly detection is an important task in quality inspection and automation.
It aims to find outliers based on the normal samples provided.
Many recent methods take a reconstructive approach that models the distribution of normal samples and then discriminates the anomalous ones.
The training of the reconstruction process often requires the presence of anomalous samples, data augmentations are therefore often used as a way to produce simulated anomalies as it is difficult to collect real samples.
In practice, the characteristics of the simulated anomalies significantly impact the quality of the reconstruction results.

The core idea of existing data augmentation methods is to randomly replace a region of normal areas with other values, thus creating an anomaly.
Two natural questions raised in this process are how to select the target region and what to use as the anomaly source.
In previous methods, the choice of target region includes rectangular areas, scars (thin rectangular areas) \cite{li2021cutpaste}, randomized masks \cite{zavrtanik2021draem}, and masks obtained by thresholding difference \cite{schluter2022natural}.
As for the anomaly source, simple CutOut (replace with zeros) \cite{devries2017improved}, random noises, external texture sources \cite{zavrtanik2021draem}, and in-distribution sampling \cite{li2021cutpaste} are widely adopted. 
As various as the choices could be, previous solutions create data augmentation methods based on two common perspectives.
They either focus on creating anomalies with high variety or approximating realistic appearances.

In this paper, we support the use of diverse shapes and randomly distributed locations in anomaly generation.
However, we oppose the empirical thoughts that believe mimicking or approximating the distribution of real anomalies could lead to optimal solutions.
The anomaly detection problem is known for the unpredictability of anomalies.
It means that we shall hold no presumptions about the appearance of anomalies in the real world.
The hypothesized anomaly distribution that previous methods try to approach is a biased distribution that is extrapolated from past observations of the test set or real-world experience.
Mimicking a biased distribution leads to biased solutions which could provide false comfort in the test set which is heavily involved with human experience in its creation process.
The results are therefore not representative enough, and the performance of these methods might not be as promising in the real world.
On the other hand, other methods seek to create anomalies with a high variety, but the current way of executing it is unsatisfying as a randomized source does not produce anomalies with a truly high variety.
Simply using randomly selected anomaly sources, whether internal or external, could only achieve variety in values, not the type of anomalies.
These attempts previous methods have made are empirical, and further analysis is required to produce anomalies with different intrinsic natures to cover the possible situations from different angles.

In terms of a reconstructive perspective, anomalies could be divided into transparent and opaque ones for they require to be treated differently by the network.
For transparent ones, the original normal areas are covered by anomalies but still visible, so the goal is to retrieve them.
For opaque ones, the original regions are completely gone, so the goal is to reconstruct based on the information from their surrounding areas.
However, the premise of this division may not hold as normality varies within a certain range in reality.
Anomalies that lie close to the normal distribution are less likely to be well reconstructed, so we separately propose a method to create near-distribution anomalies, tightening the decision boundary of normality.
Additionally, we include the rotation anomaly, which differs from previous ones as it completely originates from manual definitions in certain object classes.
Since providing samples with the same orientation during training is sufficient to learn that rotation is not allowed, and it is difficult to compute the anomaly mask created by the rotation operation, we apply this operation in reverse to classes that allow rotation to emphasize its irrelevance in creating an anomaly.

The anomaly simulation process aims for diversity, but using all augmentations indiscriminately may result in suboptimal outcomes since the same augmentation may not be considered anomalous for different classes.
Unlike previous automated methods, we believe that manual intervention is necessary to determine the appropriate combination of anomaly simulation methods since the human definition of anomalies varies in each class.
Therefore, we additionally propose a simple yet effective selection strategy that disables certain augmentations if it is considered irrelevant in creating anomalies.



We follow the previous work \cite{zavrtanik2021draem} to build a reconstructive framework to evaluate our anomaly simulation method.
A key ingredient from the previous method in tackling the overfitting issues is to use rotation augmentation on normal samples before composing anomalies.
However, it interferes with our new augmentation framework since rotation may introduce anomalies in certain classes.
We remove the arbitrary use of rotation augmentation and propose a split training strategy to improve generalization.
Specifically, we split the training data in half, and use different samples in reconstruction and localization, thus preparing the localization process for the reconstruction quality drop in practice. 
 
In experiments, our method demonstrates SOTA performance in benchmarks, and we show how different anomaly augmentations affect the final reconstruction quality on a simulated dataset.
In ablation, the effect of rotation augmentation and the splitting strategy is investigated further.
Our main contributions are listed as follows:
\begin{itemize} 
\item A comprehensive anomaly simulation framework that selectively applies different augmentations.
\item A near distribution anomaly augmentation method.
\item A split training strategy that alleviates the overfitting issue in two-stage frameworks.
\end{itemize}


\section{Related Work}
Most recent works on anomaly detection focus on unsupervised settings due to the fickle nature of anomalies and the difficulty in collecting them.
The MVTec AD \cite{bergmann2019mvtec} provides a high-quality benchmark for this problem.
Numerous methods have emerged after its release, and two major approaches are reconstruction-based methods and feature-based methods.
Self-supervised methods are often integrated into other approaches to generate simulated anomalies for training.
Here, we briefly review previous works on reconstruction-based anomaly detection and self-supervised anomaly detection to provide the necessary context to understand our work.
\subsection{Reconstruction-basd Anomaly Detection}
Typical reconstruction-based methods consist of a reconstructive network that aims to restore anomalous inputs to normal images.
Then, the anomaly level is determined by the reconstruction error between them.
Autoencoders have also been widely used for image reconstruction.
The SSIM \cite{wang2003multiscale} loss is adopted by \cite{bergmann2018improving} to aid the reconstruction process.
A U-Net \cite{ronneberger2015u} is utilized to better distinguish the anomalies from the reconstruction results \cite{zavrtanik2021draem}.
A self-supervised block \cite{ristea2022self} is proposed to serve as a ``plug and play'' component which improves the performance of many methods.
In this paper, we believe the right data stimulation is the key to better reconstructions and performance improvements in downstream tasks.
\subsection{Self-supervised Anomaly Detection}
The anomaly synthesis problem is closely connected to general data augmentation methods.
In the development of self-supervised anomaly detection, a lot of methods draw inspiration from previous general data augmentation methods.
Cutout \cite{devries2017improved} and RandomErasing \cite{zhong2020random} randomly erase a part of images and then replace it with other values to create augmentations.
CutMix \cite{yun2019cutmix} and Cutpaste \cite{li2021cutpaste} utilize image patches that are selected from themselves and paste them to other regions to create anomalies.
In DRAEM \cite{zavrtanik2021draem}, an outside data source \cite{cimpoi2014describing} is used as the anomaly source, and a Perlin \cite{perlin1985image} mask is introduced to create a mask with uncertain shapes.
NSA \cite{schluter2022natural} proposes a method that could seamlessly blend scaled patches of various sizes from separate images through integrating Poisson image editing, creating anomalies that are visually close to real-world anomalies.

The common assumption behind these methods is that the simulated anomaly should be as close to the real-world samples as possible.
We agree that the reconstruction results will be better if similar anomalies have been seen in training. 
However, we think that the presumption of real-world anomalies is highly biased as anomalies do not follow certain patterns.
Instead of trying to mimic the imaged real-world anomalies, which do not exist, this work aims to construct a framework that generates anomalies of different traits, so the reconstruction ability could later generalize to similar anomalies in inference.

\section{Method}
This paper presents a comprehensive anomaly simulation framework that aims to generate diverse anomalies while accounting for distinct standards of normality associated with each class.
The final behavior of the reconstructive network is highly related to the anomaly samples it received during training.
Since different classes do not share the same standard of normality, selectively applying different anomaly simulation methods is therefore optimal as the same augmentation may not be considered anomalous for all classes.
A split training strategy is also proposed as an alternative to alleviate the intermediate inconsistency in two-stage frameworks.

\subsection{Anomaly Simulation Framework}
Our framework consists of a set of anomaly simulation methods.
Since anomalies are known to be unpredictable both in shape and appearance, we follow the previous method \cite{zavrtanik2021draem} used to generate mask \(M_u\) with uncertain shapes.
As for appearance, we believe that relying solely on arbitrary sources is insufficient to establish a credibly diverse training distribution.
In this paper, we direct our attention toward the distinctions in the characteristics of anomalies and classify them into two categories, namely Transparent and Opaque.
The intuition behind this is that the reconstructive network performs different actions for them: restoration and reconstruction, respectively.
The original image \(I\) is covered with randomly sampled source \(N\) \cite{cimpoi2014describing}.
The transparent augmentation \(I_t\) is defined as:
\begin{align}
I_t	= \overline{M}_u \odot I + (1-\beta)(M_u \odot I)+\beta(M_u \odot N),
\end{align}
where \(\beta\) is the opacity parameter that controls the transparency, and \(\overline{M}_u\) is the inverse version of \(M_u\).
The opaque ones are generated similarly except that the beta value is fixed to one.
The opaque augmentation \(I_o\) is defined as:
\begin{align}
I_t	= \overline{M}_u \odot I + M_u \odot N.
\end{align}

\begin{figure}[tbh]
	\centering{
		\includegraphics[width=\linewidth]{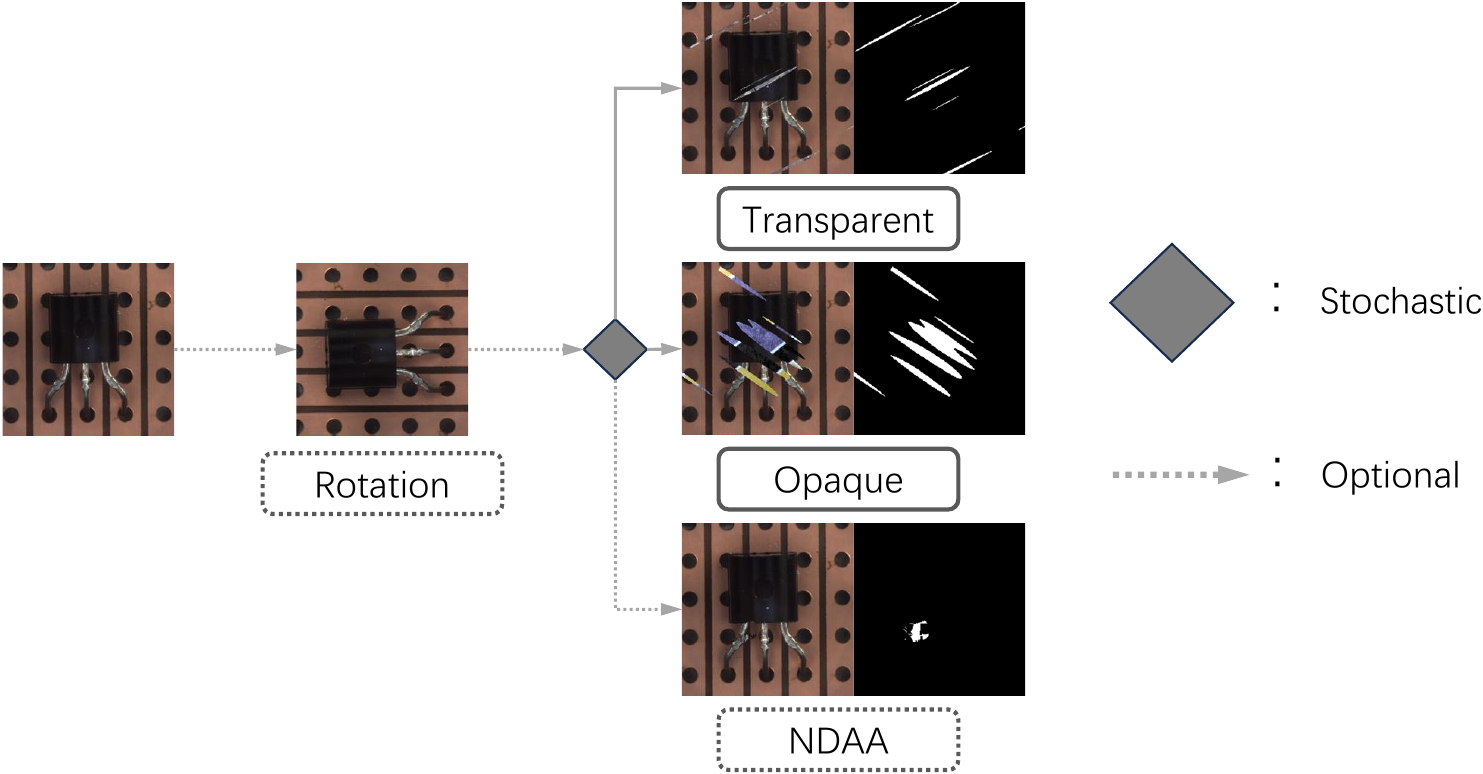}
		}
	  \caption{\label{fig: custom}
	This figure shows how to select the appropriate augmentations for training.
}
\end{figure}

If the normality is fixed, then all anomalies can be classified into the aforementioned two categories. However, the standard of normality changes within a certain range for one class, and it is different across different classes.
The global position remains relatively fixed in certain classes, thereby rendering rotations as anomalies. 
Some other classes perceive alterations in relative position as anomalies, encompassing bends and other changes in relative position.
In other classes (primarily some irregular textures), neither of these changes is a contributing factor to the occurrence of anomalies.
To tighten the decision boundary of normality, we additionally include rotations and introduce a method called Near-distribution Anomaly Augmentation (NDAA).
The aforementioned augmentation methods are different from the previous methods, and they should be judiciously employed during training to establish an optimal training distribution.
As illustrated in \cref{fig: custom}, NDAA should only be employed when changes in relative position would generate anomalies within that particular class.
If the augmented image does not consist of anomalies, then this augmentation method should be excluded.

\begin{figure}[tbh]
	\centering{
		\includegraphics[width=\linewidth]{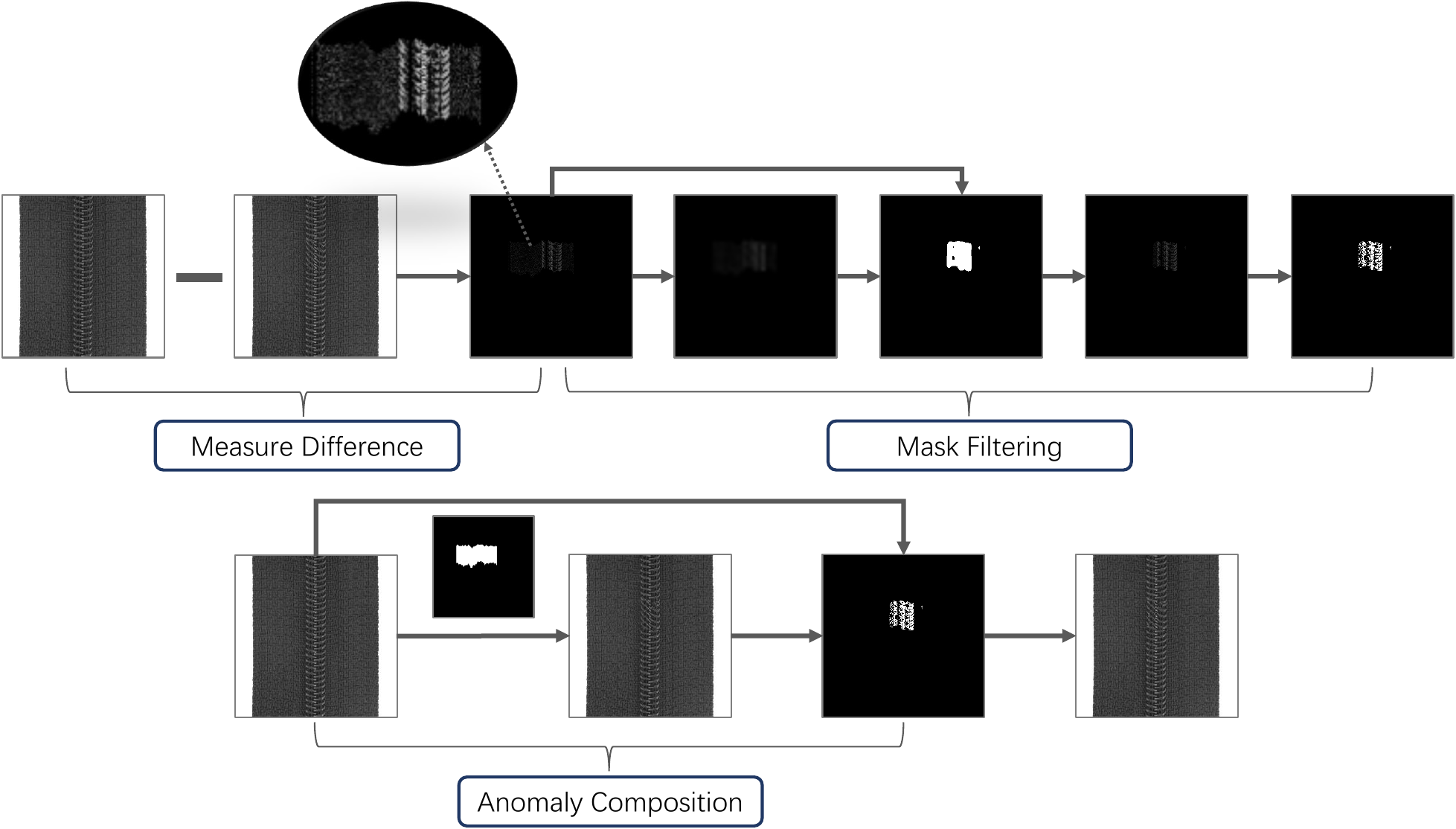}
		}
	  \caption{\label{fig: process}
	This figure illustrates the process of constructing a near-distribution anomaly.
}
\end{figure}

\subsection{Near-distribution Anomaly Augmentation}
The purpose of this method is to enhance the ability to discern anomalies that are closely distributed to the normal distribution.
The anomaly created should essentially exhibit spatial proximity to its surroundings, manifesting in various forms such as bending or distortion, while not being limited to these manifestations.

As illustrated in \cref{fig: process}, we first select a rectangular area and distort it according to a \(\sin\) curve.
Then we take the absolute difference between the original image and the distorted image and use a threshold to filter their difference into a primitive mask \(M_o\).
This process creates many scattered dots that are distributed in the background.
Therefore, we use block reduce first and then resize the reduced image back to the original size before thresholding it.
This operation connects visually near dots into contiguous regions and produces mask \(M_l\), but it also enlarges the original mask area.
We calculated a pixel-wise product between \(M_o\) and the \(M_l\), thus preserving the dense areas and filtering the noise.
The final masked regions in the original image are then replaced with the distorted area to get the final augmented image.

This simulation method is crucial in improving the reconstruction in a lot of classes because the normal samples often vary in detail.
The reconstructive network could confuse these differences with subtle anomalies, leading to non-ideal results.
The anomalies created by the NDAA method are similar to their surrounding area, thus pushing the network to discriminate the difference between them.
\begin{figure*}[tbh]
	\centering{
		\includegraphics[width=\linewidth]{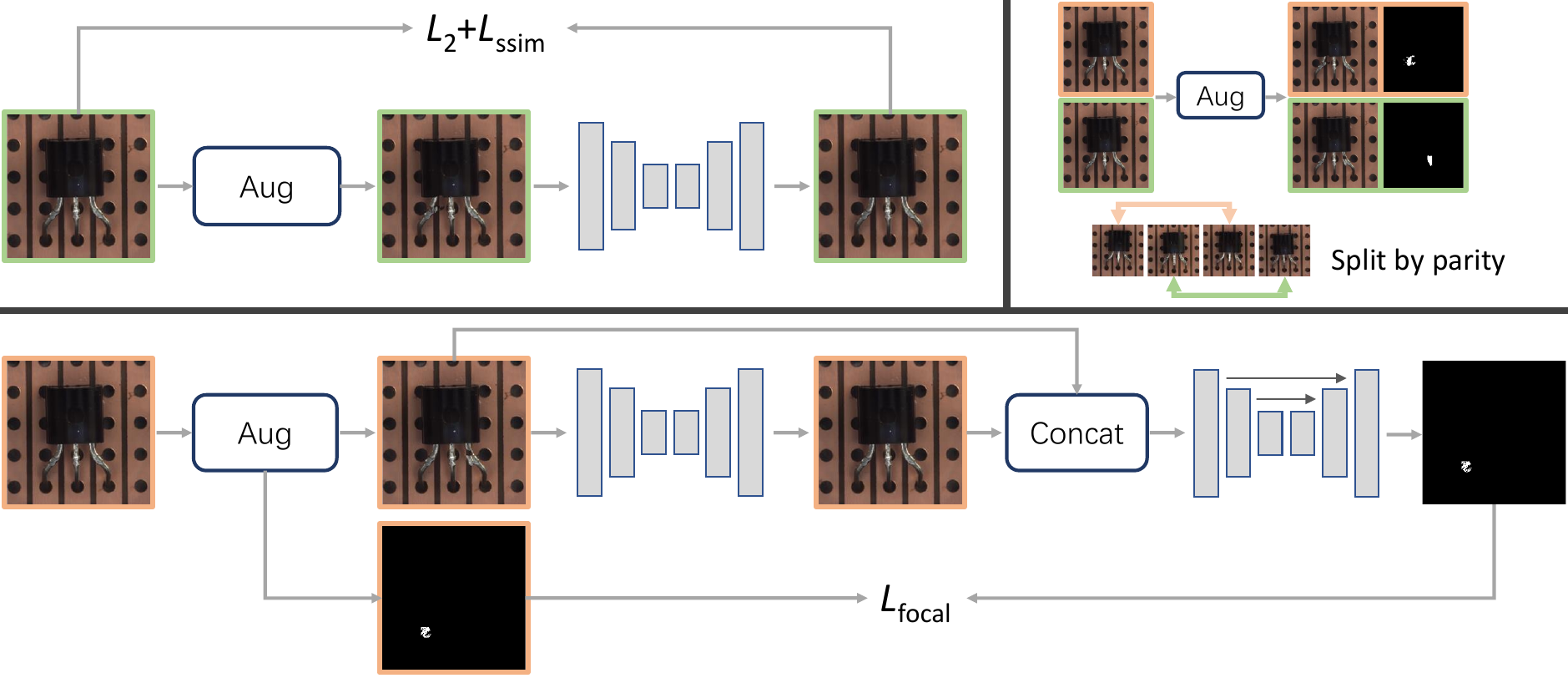}
		}
	  \caption{\label{fig: framework}
	This figure shows the structure of the reconstructive framework and the concept of the split training strategy. The green samples and the orange samples are from the two different portions of the given training distribution.
}
\end{figure*}

\subsection{Reconstructive Framework}
\label{sec: framework}
Previous star work \cite{zavrtanik2021draem} provides a simple yet powerful solution for reconstructive methods.
It comprises a reconstructive network and a discriminative network, where the former is responsible for transforming anomalous input into normal images, while the latter learns an appropriate distance function to quantify the level of anomaly based on the disparity between the input and the reconstructed image. 
In the previous work, rotation augmentation was used stochasticly to normal samples before they entered the reconstructive network.
We believe that the purpose of this operation is to address issues caused by the limited number of training samples, such as overfitting.
However, it affects the process of anomaly generation and could introduce anomalies by itself in certain cases, which leaves us concerned about its impact on the reconstruction process.
If we remove this operation, generalization problems arise immediately. 
The reconstructive network can perfectly reconstruct the samples it has seen during training, but it produces blurry results for test samples.
This inconsistency affects the downstream discriminative network because the learned distance measurement is no longer accurate due to the difference in reconstruction quality between training and testing.

Considering this circumstance, we propose a direct solution to restore downstream performance while removing the global rotation augmentation.
The discriminative network is trained with high-quality reconstruction during training, whereas the reconstructive network fails to exhibit satisfactory generalization and produces blurry reconstructions for test samples.
Therefore, we have decided to address this issue reversibly by exposing the discriminative network in training to the same reconstruction quality it would have experienced in inference.
As shown in \cref{fig: framework}, we propose a partition of the data into two disjoint subsets based on the parity of their indices.
Give a set of normal samples \(I = \{I_i : i \in \{1, 2, …, N\}\), we split it into \(I_{X} = \{I_i : i \in \{1, 2, …, N\}, i \mod 2 = 1\}\) and 
\(I_{Y} = \{I_i : i \in \{1, 2, …, N\}, i \mod 2 = 0\}\).
Then, \(I_X\) is used to train the reconstructive network, but it does not participate in the training process of the discriminative network.
\(I_Y\) is passed through the reconstructive network without calculating the reconstruction loss and the reconstruction results are provided to the reconstructive network to train it.
Following previous works, we use \(\mathcal{L}_2\) Loss as the reconstruction objective and additionally apply SSIM Loss \cite{wang2004image} to stress the interdependence in the reconstructed image.
Besides, Focal Loss \cite{lin2017focal} is used as the localization objective to focus learning on hard examples.
The full objective could be formulated as follows:
\begin{align}
\mathcal{L}(I_X,I_Y,M_{gt})	&= \mathcal{L}_{SSIM}(I_X,R(I_X))+\mathcal{L}_2(I_X-R(I_X))\nonumber\\
					&+ \mathcal{L}_{focal}(S(I_Y \oplus R(I_Y)),M_{gt}),
\end{align}
where \(\oplus\) is the channel-wise concatenation, and \(M_{gt}\) is the ground truth mask.
\(R\) and \(S\) represents the reconstructive network and the discriminative network respectively.
\(\mathcal{L}_{SSIM}\) is a patch based SSIM loss, and \(\mathcal{L}_{focal}\) refers to Focal Loss.

In this way, the reconstruction quality remains consistent in the samples encountered by the discriminative network during both training and testing, leading to stable performance in the downstream task. 
Different ratios of these two portions have been experimented with, and empirical evidence suggests that assigning equal sample sizes by parity produces more favorable outcomes.

\section{Experiments}
The performance of our method is compared with previous methods, and empirical evaluations demonstrate the quality improvements brought by our anomaly simulation method in reconstruction results
We additionally assess the generalization performance by cross-comparing the performance of each method on a simulated dataset.
\subsection{Benchmarks}
In this section, we provide the benchmarks comparing our performance with previous methods (DRAEM \cite{zavrtanik2021draem}, NSA \cite{roth2022towards}, PatchCore\cite{roth2022towards}).

{\bf Dataset}
Experiments in this paper are conducted on the MVTec \cite{bergmann2019mvtec} anomaly detection dataset.
The MVTec dataset contains 15 classes including 5 classes of textures and 10 classes of objects.
This dataset provides a training set with only normal images and a test set comprised of various anomalies.
It provides pixel-level annotations which allow benchmarks for anomaly localization.

{\bf Experimental settings}
All the images are resized to a size of \(256 \times 256\) before entering the network.
The training settings and the model choices mostly follow the previous work \cite{zavrtanik2021draem} to make a fair comparison, as this paper mainly focuses on improving performance through more comprehensive training data.
We randomly split the training data in half and used them for training each network separately.
The data collection process could store similar samples in near positions, so it is worth noting that the data is separated in parity order instead of upper and lower halves.
Also, there is no indiscriminate use of image rotation (on anomaly-free images as a data augmentation method, not to simulate anomalies) to alleviate the overfitting issue.

{\bf Metrics}
The norm in benchmarking anomaly detection methods is to report the AUROC score.
Since the anomaly detection problem is severely imbalanced and AUROC may produce a less representative score \cite{vskvara2023auc}, we additionally report the AP score to provide more representative benchmarks.

\begin{table}[tbh]
\centering
\small
\begin{tabularx}{\linewidth}{@{}l *4{>{\centering\arraybackslash}X}@{}}
\toprule
Class& DRAEM&NSA&PatchCore&	Ours\\
\midrule
\midrule
capsule&95.5 / 99.1&93.7 / 98.7&\bf97.1 / \bf 99.2 & 95.5 / 99.1\\
bottle&96.7 / 98.4&97.6 / 99.0&\bf100 / \bf100&96.5 / 98.2\\
carpet&94.7 / 98.4&85.5 / 94.5&97.9 / 99.4&\bf99.0 / \bf99.7\\
leather&\bf 100 / 100&\bf 100 / 100&\bf100 / 100&\bf100 / 100\\
pill&97.2 / 99.5&98.4 / 99.7&93.8 / 98.8&\bf98.7 / \bf99.8\\
tran&90.8 / 89.2&94.2 / 93.1&\bf100 / 100&{\bf 100} / 99.9\\
tile&\bf100 / 100&\bf100 / 100&98.7 / 99.6&\bf100 / 100\\
cable&92.9 / 96.0&94.6 / 94.0&\bf98.5 / \bf 99.0&92.5 / 95.8\\
zipper&\bf100 / 100&\bf100 / 100&99.6 / 99.9&\bf100 / 100\\
toothbrush&\bf100 / 100&\bf100 / 100&\bf100 / 100&\bf100 / 100\\
metal nut&\bf99.7 / \bf99.9&95.5 / 99.6&\bf99.7 / \bf99.9&98.7 / 99.7\\
hazelnut&\bf100 / 100&95.4 / 97.3&\bf100 / 100&98.7 / 99.2\\
screw&{\bf 97.3} / 96.7&88.6 / 96.3&97.2 / \bf 98.9&95.7 / 98.6\\
grid&\bf100 / \bf100&99.5 / 99.8&97.0 / 99.0&99.5 / 99.8\\
wood&99.6 / 99.9&96.6 / 98.9&99.4 / 99.8&\bf100 / \bf100\\
\midrule
avg&97.6 / 98.5&96.2 / 98.0&\bf98.6 / \bf99.5&98.3 / 99.3\\
\bottomrule
\end{tabularx}
\caption{
Anomaly detection results (AUROC/AP).
}
\label{tab: detection}
\end{table}%


\begin{table}[tbh]
\centering
\small
\begin{tabularx}{\linewidth}{@{}l *4{>{\centering\arraybackslash}X}@{}}
\toprule
Class&	 DRAEM&NSA	&PatchCore	&Ours\\
\midrule
\midrule
capsule&94.3 / 49.4&97.6 / \bf 55.5&{\bf98.7} / 45.5&94.6 / 41.9\\
bottle&\bf 99.1 / 86.5&98.3 / 82.0&97.9 / 76.5&98.7 / 86.2\\
carpet&95.5 / 53.5&90.5 / 36.2&98.6 / 59.4&\bf 99.3 / 82.4\\
leather&95.6 / \bf75.3&{\bf99.5} / 59.0&98.8 / 41.5&99.1 / 74.7\\
pill&97.6 / 48.5&{\bf 98.1} / 71.0&97.3 / \bf74.9&96.9 / 41.8\\
transistor&90.9 / 50.7&84.8 / 49.5&\bf 96.6 / 69.4&93.1 / 56.9\\
tile&99.2  /  92.3&99.3 / 93.2&94.7 / 50.7&\bf 99.6 / 96.8\\
cable&94.7 / 52.4&87.2 / 29.5&97.9 / 64.9&\bf 97.9 / 72.4\\
zipper&98.8 / \bf81.5 &94.2 / 67.8&97.9 / 52.8&{\bf99.0} / 66.5\\
toothbrush&98.1 / 44.7&92.9 / 40.5&\bf 98.6 / 56.6&98.3 / 42.3\\
metal nut&\bf 99.5 / 96.3&98.3 / 93.5&98.4 / 90.3&99.1 / 93.5\\
hazelnut&\bf 99.7 / 92.9&97.6 / 55.2&98.4 / 56.9&99.7 / 92.5\\
screw&97.6 / 58.2&96.1 / 42.3&99.0 / 35.9&\bf 99.4 / 68.2\\
grid&\bf 99.7 / 65.7&99.1 / 51.2&97.9 / 32.1&99.5 / 64.9\\
wood&96.4 / 77.7&91.1 / 55.6&93.0 / 46.6&{\bf 96.7} / \bf 82.3\\
\midrule
avg&97.3 / 68.4&95.0 / 58.8&97.6 / 56.9&\bf 98.0 / 70.9\\
\bottomrule
\end{tabularx}
\caption{
Anomaly localization results (AUROC/AP).
}
\label{tab: localization}
\end{table}%

{\bf Experimental results}
The results of image-level anomaly detection on MVTec are presented in \cref{tab: detection}. 
Our method demonstrates comparable performance in this task and achieves the highest scores in seven classes.
The average performance of our methods surpasses that of DRAEM and NSA, achieving an AUROC and AP of 98.3\% and 99.3\% respectively, which is slightly lower by 0.3\% and 0.2\% compared to PatchCore.
In \cref{tab: localization}, our method surpasses the state-of-the-art by 2.5\% in terms of AP and delivers a slightly better result in AUROC on the anomaly localization task.
Further inspection shows that the increase in performance is mainly because of the improvements in the reconstruction quality.
Despite achieving a better score in seven classes, our method performs less optimally in other classes.
After investigating the test set, we believe it may be attributed to the limited number of anomaly categories and inaccurate labels.
For example, the capsule in \cref{fig: quality} is squeezed, resulting in a thinner middle section compared to the right portion both above and below.
However, only the missing region at the top is identified as an anomaly.
Given that, we believe that the standard benchmark only is insufficient, and provides closer inspections on the reconstruction quality.
A simulated dataset is also constructed to evaluate our method from another perspective.

\subsection{Comparision on Reconstruction Quality }
The core idea of this work is to improve the reconstruction quality by providing more comprehensive simulated anomalies, thus benefiting the downstream detection and localization tasks.
More reconstructed anomalous areas allow the discriminative network to produce more accurate results.

\begin{figure}[tbh]
	\centering{
		\includegraphics[width=\linewidth]{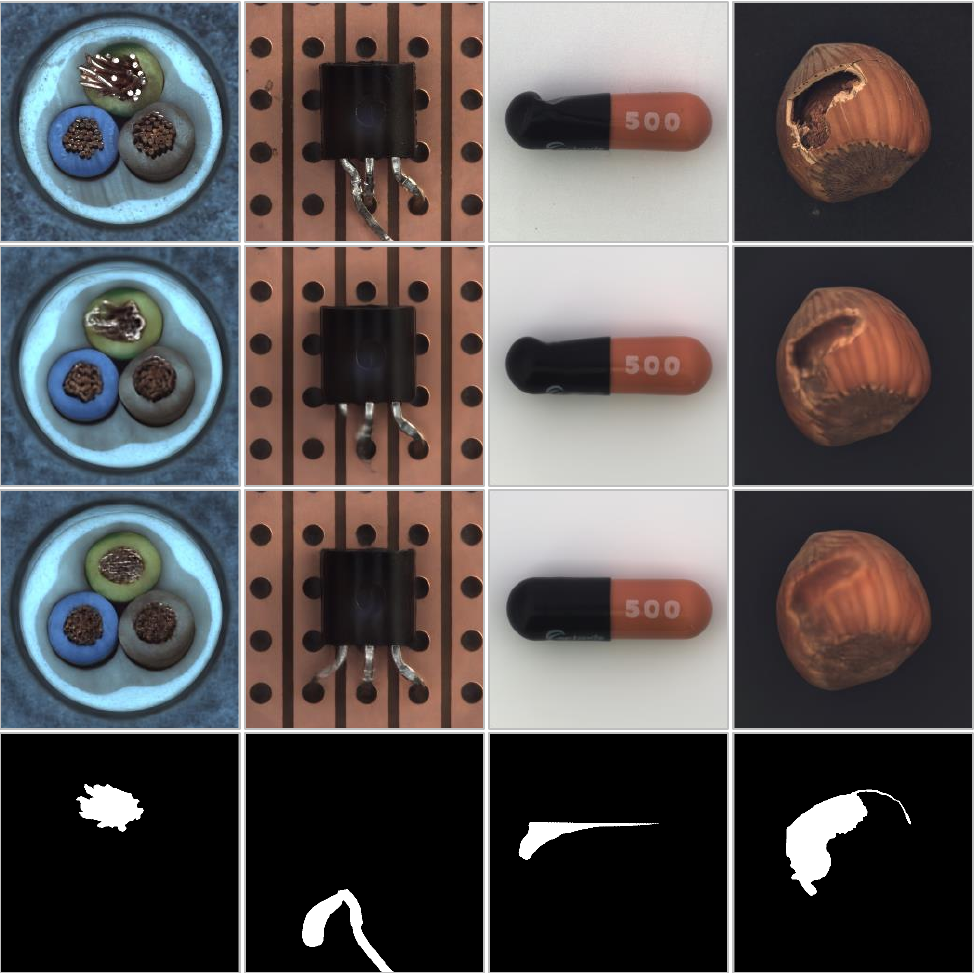}
		}
	  \caption{\label{fig: quality}
	This figure compares the reconstruction quality between DRAEM (second row) and ours (third row).
}
\end{figure}

We inspect and provide qualitative results comparing the reconstruction quality of hard samples between our method and the previous method \cite{zavrtanik2021draem} while using an identical reconstructive network. 
Qualitative results are provided in \cref{fig: quality}.
The first column displays an anomaly that arises from the distortion of normal samples, wherein the metal structure at the top is tilted.
The metal structure in its normal form could vary within a certain range, thus making it hard to distinguish between an anomaly and a normal variant.
The previous approach encounters difficulties in detecting such anomalies, which we attribute to its training on simplistic simulated anomalies that do not encompass the scenarios of in-distribution anomaly sources.
The proposed method is trained using a wider range of anomaly categories, and a selected combination of these categories is tailored to match the characteristics of specific classes. 
As a result, our method exhibits enhanced capability in accurately detecting anomalies that are closely distributed among normal samples.
The second column shows a different situation, where the bent metal strip also creates a missing part on the left.
The previous method is capable of effectively removing redundant components, but it fails to sufficiently recover missing components.
At a local level, the missing part leaves a background that appears the same as other normal regions, making it hard to distinguish whether it should be recovered. 
The proposed method effectively eliminates the deformed metal strip and precisely restores it to its original position, showcasing its superior capability in accurately modeling global normality.

The utilization of a more comprehensive simulated training dataset enables our model to possess a higher likelihood of achieving ideal reconstructions due to its prior exposure to anomalies exhibiting similar characteristics during training.
The effectiveness of our proposed anomaly simulation framework in enhancing reconstruction quality is demonstrated, by utilizing a minimal structure comprising solely a reconstructive network and the corresponding anomaly simulation method.
Therefore, it could presumably be integrated into other reconstruction-based methods that utilize simulated anomalies to further improve their reconstruction quality and subsequently increase downstream performance.

\subsection{Generalization on Simulated Anomalies}
\label{sec: sim}
In this study, we posit that the performance of the reconstructive network on unseen anomalies can be enhanced through training with simulated anomalies exhibiting similar characteristics.
We argue that previous methods, which aim to replicate the distribution of actual anomalies, introduce an inductive bias into the network. The empirical conclusion of the distribution of real anomalies is based on our prior experience with the test set. 
Additionally, new datasets for testing are also created based on these observations. 
In essence, there could be a significant overlap between the modeled real-world distribution and the test set since both are generated under our assumptions regarding real-world anomalies.
Besides, the test sets in current datasets only contain anomalies of limited types, which is concerning since the evaluation results could be presumably beneficial to certain methods. 

\begin{table}[tbh]
\centering
\small
\begin{tabularx}{\linewidth}{@{}l *4{>{\centering\arraybackslash}X}@{}}
\toprule
Class&DRAEM&NSA&Patchcore&Ours\\
\midrule
\midrule
Cutpaste&86.8 / 58.1& 82.2 / 30.2&{\bf  93.5} / 52.8& 89.3 / \bf 60.4\\
NDAA&96.4 / 67.2  & 91.7 / 31.6&95.0 / 30.7&\bf 96.9 / 72.3 \\
NSA&97.9 / 82.6& \bf 98.4 / 92.7 & 94.7 / 66.4& 96.6 / 82.8\\
Opaque&\bf 100 / 99.5& 87.0/ 49.2& 94.0 / 45.4& 99.9 / 98.8 \\
Transpa-&\bf 100 / 99.9& 86.5 / 43.2 &95.5 / 48.1& 100 / 99.3\\
Avg& 96.2 / 81.4&89.1 / 49.3& 94.5 / 48.6&\bf 96.5 / 82.7\\
\bottomrule
\end{tabularx}
\caption{
Anomaly localization results on the simulated dataset.
}
\label{tab: gen}
\end{table}%

Given this, the test set might not be representative enough to reflect the performance in real-world scenarios truthfully.
However, it is currently infeasible for researchers to create a dataset with a truly representative test set, considering the massive cost and rarity of anomalies.
Therefore, based on the definition we made previously on the categories of anomalies, we proposed to utilize these synthetic methods to create a simulated dataset to conduct further evaluations of the models.
Although we could not guarantee that the methods that perform well in the simulated dataset will generalize well in real-world scenarios, it is however certain that a method that performs poorly under simulated scenarios will be harder to claim good generalization performance in the real world.


\begin{table}[tbh]
\centering
\small
\begin{tabularx}{\linewidth}{@{}l *4{>{\centering\arraybackslash}X}@{}}
\toprule
Class&D.NoRot&D.Our&Our.D&Ours\\
\midrule
\midrule
capsule&	93.9 	/	47.3 	&	93.0 	/	37.2 	&	89.5 	/	51.8 	&	94.6 	/	41.9 	\\
bottle&	98.3 	/	86.5 	&	98.7 	/	85.9 	&	98.9 	/	87.3 	&	98.7 	/	86.2 	\\
carpet	&	92.9 	/	28.0 	&	98.8 	/	72.2 	&	93.6 	/	50.0 	&	99.3 	/	82.4 	\\
leather	&	98.0 	/	69.8 	&	98.1 / 65.7 	&	99.1 	/	74.7 	&	99.1 	/	74.7 	\\
pill	&	94.6 	/	50.0 	&	96.1 	/	33.1 	&	97.1 	/	41.4 	&	96.9 	/	41.8 	\\
transistor	&	82.6 	/	39.5 	&	85.4 	/	33.0 	&	87.9 	/	46.7 	&	93.1 	/	56.9 	\\
tile	&	97.6 	/	86.5 	&	99.4 	/	96.6 	&	99.6 	/	96.8 	&	99.6 	/	96.8 	\\
cable	&	92.6 	/	52.1 	&	92.9 	/	51.4 	&	94.8 	/	59.0 	&	97.9 	/	72.4 	\\
zipper	&	95.2 	/	47.5 	&	81.6 	/	14.4 	&	97.1 	/	72.5 	&	99.0 	/	66.5 	\\
toothbrush	&	98.4 	/	58.5 	&	96.0 	/	26.1 	&	97.7 	/	52.0 	&	98.3 	/	42.3 	\\
metal nut	&	96.0 	/	85.1 	&	97.9 	/	81.4 	&	99.3 	/	94.6 	&	99.1 	/	93.5 	\\
hazelnut	&	99.3 	/	82.6 	&	99.7 	/	95.0 	&	98.7 	/	78.6 	&	99.7 	/	92.5 	\\
screw	&	98.7 	/	41.2 	&	99.4 	/	66.5 	&	98.8 	/	65.6 	&	99.4 	/	68.2 	\\
grid	&	99.5 	/	63.5 	&	99.6 	/	66.0 	&	99.5 	/	55.8 	&	99.5 	/	64.9 	\\
wood	&	84.7 	/	42.8 	&	95.6 	/	68.2 	&	96.7 	/	76.5 	&	96.7 	/	82.3	\\
\midrule
avg	&	94.8 	/	58.7 	&	95.4 	/ 59.5 &	96.6 	/	66.9 	&	98.0 	/	70.9 	\\

\bottomrule
\end{tabularx}
\caption{
Anomaly localization results of the ablation study. From left to right, the listed methods are DRAEM without using rotation augmentation, DRAEM using our simulation method, our architecture using the simulation method of DRAEM, and our original method.
}
\label{tab: ablantion}
\end{table}%

Specifically, the simulated dataset is constructed by applying the proposed anomaly simulation methods on the anomaly-free images from the test set of the Mvtec AD \cite{bergmann2019mvtec}.
Since our method is trained with these anomalies, we could not eliminate the possibility that our method performs better because the anomalies are generated with the same method.
Therefore, the anomaly simulation methods proposed by previous works are additionally included in the simulated dataset for a fair comparison.
Each category contains simulated anomalies and their corresponding normal samples, and we used the same model trained in the previous section to conduct evaluations.
By introducing other simulation methods, we show how our method reacts to anomalies it has not seen in training, verifying that the anomaly simulation method is comprehensive and its categories are inclusive and enable the network to generalize to anomalies of unseen appearance.

The results of the anomaly localization task on the simulated data are presented in \cref{tab: gen}, and our method demonstrates superior performance on average.
We have observed that our method has not only performed well on the simulated anomalies proposed in this paper but also achieved competitive results in the other two categories which are generated from methods proposed by previous studies.
Besides, we could observe that the models exhibit a higher performance within the corresponding category utilized during training.
The findings validate our hypothesis that models will exhibit superior performance when dealing with anomalies belonging to the same category as those encountered during training.

\subsection{Ablation Study}
In this work, the indiscriminate use of rotation augmentation is removed since it contradicts the core design philosophy of our anomaly simulation framework.
We report on ablations for the removal of the rotation augmentation and the choice of the anomaly simulation framework.

{\bf Rotation augmentation}
The rotation augmentation and the split training strategy are both solutions proposed to enable better generalization.
We believe that the indiscriminate use of rotation augmentation in the anomaly detection task is undesirable as it can alter the characteristics of simulated data and introduce anomalies in inputs, significantly impacting reconstruction quality in certain classes.

The removal of it, as demonstrated in \cref{tab: ablantion}, however, results in severe overfitting issues and a significant decrease in performance.
Further inspection reveals that a lot of noises emerge in the anomaly-free area.
The reconstructive model overfits and produces perfect reconstruction in training, while producing less satisfying reconstructions in inference, resulting in a quality gap.
The discriminative network failed to adapt to this gap, and it started classifying normal areas with less accurate reconstructions as anomalies because it was only exposed to perfectly reconstructed samples during training.
Therefore, we develop the split training strategy as an alternative for rotation augmentation while not interfering with the characteristics of the simulated data.

{\bf Anomaly simulation methods}
To validate the effectiveness of the proposed anomaly simulation framework, we cross benchmark the results of using different architecture and data simulation methods and report the results in \cref{tab: ablantion}.
We could observe that directly combining the DRAEM architecture with our simulation methods does not yield satisfying results.
The first cause is that this combination is trained without both the rotation augmentation and the spitting training strategy, which leads to overfitting.
Although NDAA helps the reconstructive network better model the variation range of normal samples, we think that, empirically, it does increase the risk of overfitting.
The lack of methods to mitigate it makes this issue even more significant in this particular context.
If we reversely use our new architecture with the anomaly simulation method of DRAEM, the results are only slightly different from the original DRAEM model, which is expected.
The utilization of the split training strategy becomes less necessary, as the rotation augmentation in the anomaly simulation method of DRAEM effectively addresses overfitting concerns, therefore an integrated training scheme is potentially more stable in this case.

\section{Conclusion}
This paper introduces a comprehensive anomaly simulation framework, comprising four distinct anomaly simulation methods and a selective strategy for determining the appropriate combination of simulated anomalies.
A reconstructive framework trained under a split training policy is developed to incorporate the anomaly simulation framework while utilizing its strength to serve the anomaly detection task.
In experiments, the proposed method achieves a new state-of-the-art on the MVTec anomaly detection dataset by an AUROC of 98.0\% and an AP of 70.9\% on the anomaly localization task.
Further experiments demonstrate the leading cause of the performance improvements is better reconstruction quality brought by a more comprehensive anomaly simulation framework.
To enhance the representativeness of the results, a simulated anomaly dataset that contains anomalies of various kinds is created, and the benchmarks further show our method has more potential to excel against various unknown anomalies in the real world.
 
\section*{Acknowledgments}
We thank anomalib \cite{anomalib} for the code support.
This work was supported by the National Natural Science Foundation of China under grant 62201142.

\bibliography{aaai24}

\end{document}